# Depression Detection Using Digital Traces on Social Media: A Knowledge-aware Deep Learning Approach

Wenli Zhang, Ph.D.
Iowa State University
3332 Gerdin, 2167 Union Drive
Ames, IA, USA 50011-2027
Email: wlzhang@iastate.edu

Jiaheng Xie, Ph.D.
University of Delaware
217 Purnell Hall
Newark, DE, USA 19716
Email: jxie@udel.edu

Xiang Liu
University of Delaware
303 Alfred Lerner Hall
Newark, DE, USA 19716
Email: dennisl@udel.edu

Zhu Zhang, Ph.D.
University of Rhode Island
7 Lippitt Road Ballentine Hall
Kingston, RI, USA 02881
Email: zhuzhang@uri.edu

Please send comments to Wenli Zhang at wlzhang@iastate.edu.

# Depression Detection Using Digital Traces on Social Media:
# A Knowledge-aware Deep Learning Approach

**Abstract:** Depression is a pressing yet underdiagnosed issue in health management. Because depressed patients share their symptoms, life events, and treatments on digital platforms, IS scholars resort to user-generated digital traces for depression detection. While they facilitate innovative IT approaches to alleviate the social and economic burden of depression, most studies lack effective means to incorporate domain knowledge in depression detection systems or suffer from feature extraction difficulties. Following the design science research in IS, we propose a Deep-Knowledge-aware Depression Detection system to detect social media users at risk of depression and explain the detection factors. We deploy extensive empirical analyses to evaluate our designed IT artifact, which shows domain knowledge greatly improves performance. Our work has significant implications for IS research in knowledge-aware machine learning, digital traces utilization, and generalizable design principles. Practically, the early detection and factor explanation from our IT artifact can assist depression management and enable large-scale assessment of the population's mental health.

## 1. Introduction

The Fourth Industrial Revolution is characterized by technological advancements in high-speed Internet, artificial intelligence, big data analytics, and cloud computing [65]. Along with its emergence and progress, numerous IT artifacts (e.g., sensors, tools, and information systems) capable of producing or collecting data are implemented by organizations worldwide to modernize their services, scale their business, or improve the efficiency of data exchange. Meanwhile, these IT artifacts record massive amounts of data describing the context and outcomes of users' actions and forming each user's unique digital traces [30]. As part of this trend, the digital traces of user-generated content on social media represent a massive and novel source of ecological data that describes human behaviors and psychological characteristics [13,25,47,68] which present considerable research opportunities for Information Systems (IS) researchers to evaluate and refine theories, as well as design innovative artifacts to address social or healthcare issues [26,84].

In the 21st century, depression is one of the major contributors to the overall global disease burden[1]. An estimated 3.8 percent of the world's population suffers from depression [77]. In the United States (US), more than 21 million adults have had depression, representing 8.4% of all US adults [54], and approximately $225 billion is spent on depression-related treatments and services every year [55]. Meanwhile, depression is also the primary cause of lost productivity, with employers in the US losing more than $23 billion each year as a result of its effects on absenteeism and presenteeism[2] [77]. Although considerable effort has been devoted to this research area, depression remains difficult to detect [77]. First, there is no reliable laboratory test for diagnosing depression; the diagnoses are normally based on patients' self-reported experiences. Second, current electronic health record (EHR) systems lack the tracking of behavioral data for effective depression detection. Third, people with depression have depressive and non-depressive episodes which further complicate accurate diagnosis. Meanwhile, depression continues to be underdiagnosed [59]. First, myths and misunderstandings cause individuals and primary care physicians to be unaware of depression symptoms. Second, the stigma associated with depression causes people to hide the issues and delay help-seeking. Third, depressed people may be unable to access healthcare services due to a lack of transportation, financial resources, or insurance.

Improving approaches for detecting and surveilling depression is key to combating depression and has far-reaching health and societal implications. This study focuses on detecting depressive symptoms and detectable risk factors rather than clinically diagnosing patients, as this is cost-efficient while still assisting interventions. The rapid proliferation of social media and the associated social media users' digital traces open up exciting possibilities in depression detection. A diverse set of quantifiable signals relevant to depression, including symptoms, major life events, and treatments are observed on social media [19,53]. Research suggests that health digital traces on social media provide a great means for capturing an individual's current states of mind, feelings, behaviors, and activities that often characterize depression [17,53]. Existing research shows that social media-based depression screening has the potential to achieve detection results that are comparable to unaided clinician assessment and screening surveys [29]. Meanwhile, unlike traditional survey- or interview-based depression screening approaches, which take a snapshot of a patient's mental health conditions, social media-based techniques continuously collect digital traces that allow the tracking of an individual's mental health over time. Moreover, digital traces enable non-reactive research, in which data records are not generated through contact between researchers and subjects, logging "naturally" occurring evidence of social behavior or psychological traits [3].

[1] A concept developed by the Harvard School of Public Health, WHO, and the World Bank, which is used to calculate disability, premature death, and other factors.
[2] When employees are present for work but less productive due to their illness.

Social media-based depression screening methods, using digital traces generated by social media users, can facilitate innovative IT approaches to fight depression and alleviate its social and economic burden. For social media users, such methods can provide early detection and raise awareness for those at risk of clinical depression. For social media platforms, depression screening techniques enable them to develop new services with personalized recommendations for users with depression (e.g., encourage them to seek help and treatments; promote educational content and tools, treatment options, social support, etc.) and foster supportive environments for patients suffering from depression. For public administration, early detection of depression using readily available patient-generated digital traces can direct resources to areas of high incidence and services most needed, improving the effectiveness of mental health programs. For policymaking, surveilling large-scale patient-generated digital traces can help create evidence-based policies tailored to the specific needs of different groups.

While promising, the current mainstream approaches for detecting depression on social media have major limitations. Most of these methods use feature engineering for depression detection.

However, the adopted features, including LIWC, n-grams, and sentiment, are not clinically relevant, therefore deteriorating detection performance [13,29]. With the development of deep learning, an increasing number of researchers turn to end-to-end sequence models for depression detection using raw social media posts [38,43,49]. However, limitation persists due to depressive disorders largely displaying subtle and implicit changes in language and behavior, such as a switch in the types of topics [20]. Such patterns are difficult to model unless massive labeled data is available for training deep learning models. A small group of researchers rely on social media users' online behaviors: e.g., the engagement of social media usage, the changes in egocentric networks [17], and dictionary-based symptom and drug taggers [33]. However, these predefined features do not align with the medical knowledge in depression diagnosis and do not significantly improve detection performance. Moreover, the symptomatic episodes of depression come and go. Patients do not exhibit depressive symptoms consistently on social media. The existing methods lack effective means of capturing the dynamic patterns that characterize depression.

To cope with the shortcomings of the previous approaches, guided by medical knowledge in depression diagnosis, we devise a Deep Knowledge-aware Depression Detection (DKDD) framework using digital trace data on social media. DKDD identifies depression diagnoses-related entities which are clinically relevant, incorporates medical knowledge in the model, and considers the dynamic pattern of depression. Our work has important implications for IS research. Positioned in the computational design science research, this study proposes a knowledge-aware machine learning method that sheds light on other domain-knowledge-rich areas in IS, such as healthcare, finance, and legal research, in which domain knowledge can be meaningfully incorporated into machine learning to develop novel and impactful artifacts [1,13,83]. While other knowledge-aware machine learning models exist, our study explores a proper mechanism to incorporate ontology knowledge, which is most suited to other research with known ontologies. The proposed system is our major contribution to the IS knowledge base [56]. This system offers generalized design principles for design science research in IS: integrating domain knowledge in predictive model design improves performance, and ontology is an appropriate knowledge structure. Our work also has significant practical implications for depression management. It provides an accurate detection method that can provide complementary information to existing depression screening procedures. For public health management, our method enables large-scale analyses of a population's mental health beyond what has previously been possible with traditional methods.

## 2. Related Work

In this section, we review four bodies of literature related to our research. We first position our work in the health IS research and identify the research opportunity. Next, we conduct an extensive review of the existing social media-based depression detection studies and discover the research gap. To fill such a gap, we summarize knowledge-aware machine learning in the next subsection. Finally, we review the machine learning techniques related to our system design.

### 2.1. Healthcare Predictive Analytics in IS

Healthcare predictive analytics is an indispensable IS area that examines patient data and makes diagnostic or prognostic risk predictions [66]. Its applications range from identifying patients at risk [8], supporting clinical decision-making [10], to dealing with hospital administrative challenges [51]. Our study belongs to healthcare predictive analytics in IS: we address the difficulties of accurate depression detection using digital trace data generated by social media users.

Digital traces have significant potential and implications for addressing socio-technical concerns. Google is among the first organizations to conduct groundbreaking research using digital trace data (i.e., search queries) to detect influenza epidemics [26]. Soon, leveraging digital trace data becomes an emerging and vital research avenue in IS [4,30]. IS healthcare analytics researchers have used digital trace data to generate new data science artifacts to address healthcare challenges. For example, [84] extract digital traces from social media, environmental sensors, and healthcare records to identify asthma risk factors. [80] retrieve patients' health digital traces to predict hospital readmissions. While academics have suggested that unique digital trace data combined with domain knowledge and appropriate analysis techniques can serve as the foundation for a "21$^{st}$-century science" [30,75], they have also pointed out that IS researchers need new tools to leverage digital trace data and exploit them effectively [30]. Our work aligns with this research direction: we aim to design a new artifact that elevates medical knowledge to decipher digital trace data and identify patients at risk of depression.

### 2.2. Depression Detection Using Digital Traces on Social Media

Social media has been extensively studied by IS and is seen as having tremendous potential in various IS regimes [2,13,15,23,25,28,35,40,42,47,48,57,68,71,72,81]. According to the uses and gratifications theory, the gratifications of using social media include expression of opinion, social interaction, information sharing or seeking, etc [76]. Applying the network externalities and motivation theory, [44] show that enjoyment, peers, and usefulness explain why people continue to use social media. Especially, depressed patients are motivated to share their symptoms, major life events, and treatments for offering or seeking support and fighting the stigma of mental illness [18], examples of which are reported in Table 1. It is worth noting that there are other traces for detecting depression, such as family history, genetics, and poor nutrition. However, since our research context revolves around social media, we concentrate on factors that are detectable on these platforms. These include self-reported symptoms and major life events that may contribute to depressive symptoms, and individuals' self-reports of depression, including their use of antidepressants and treatments. Research on social media for depression analyses focuses on two categories: (1) the correlations between social media use and mental illnesses [37], and (2) using social media data for mental disorder detection [29]. Our work focuses on the latter. [Insert Table 1 here]

Most existing studies use sentiment analysis and lexicon-based features for depression detection (see Table 2). For example, [13] conduct LIWC and sentimental analyses, followed by rule-based classification for finding people with emotional distress. Although practicable, there is a significant discrepancy between the features used in these studies and medical practitioners' criteria in depression detection. Most people experience low sentiments occasionally and certain individuals tend to favor specific words depending on their education level, the influence of peers, and social context. Sentiment analysis and lexicon-based features may provide insights into one's psychological states, but they do not specify depression. [Insert Table 2 here]

A few prior studies consider entities in social media data that can characterize depression. One of the first studies is by [20], in which the authors identify social media users' behavioral patterns, including social engagement and exercises. Similarly, [21] examine social media users' behavior attributes, such as engagement in social media, the change in egocentric social networks, emotional states, and depression-related topics. Although interesting and promising, the attributes used in these two studies still considerably differ from the medical definition of depression. Depression has received concerted attention from many practitioners and researchers. There is established medical knowledge in depression detection and diagnosis, such as [5,9,50,64]. *Our first research question (R1) is how to leverage medical domain knowledge for social media users' depression detection*.

In 2020, [33] pushes this research area forward by proposing depression marker taggers to identify depression-related symptoms and drug-use experiences. This work is the closest to ours in motivation. However, this work and other existing studies use predetermined dictionaries and hand-crafted features, which limit their methods to find novel (e.g., model's unseen features) and significant features for depression detection. Hence, we are motivated to develop a model that can automatically extract entities in social media posts directly related to clinical depression diagnoses; simultaneously, these entities are supported by existing literature as being self-reported by social media users, including depression symptoms, life events that may cause or exacerbate depression, and depression treatments [9]. *Our second research question (R2) is how to effectively extract depression diagnosis-related entities*.

Furthermore, depressed patients have depressive episodes, which are periods characterized by low mood and other depression symptoms that last for two weeks or more [9]. Depressive episodes may occur from time to time. The existing methods do not capture the varied importance of digital traces occurring at different time points. In the meantime, according to medical domain knowledge, different depressive symptoms indicate varying degrees of severity; different depression diagnosis factors have different effects on the onset and exacerbation of depression. Such medical knowledge has been largely neglected by most existing studies on social media-based depression detection. *Our third research question (R3) is how to incorporate knowledge (i.e., recency, frequency, and relevancy to the onset of depression) associated with depression diagnosis-related entities for depression detection*.

### 2.3. Knowledge-aware Machine Learning

Researchers in various fields have demonstrated that knowledge-aware machine learning, in which domain expertise or domain knowledge is explicitly and meaningfully incorporated in the design of machine learning models, can play an important role in many applications involving difficult learning tasks and limited training resources. This is because knowledge-aware machine learning models have clear advantages in streamlining model architectures, lowering training costs, and increasing model interpretability [16,63]. Knowledge-aware machine learning can be generally divided into two categories: knowledge discovery and knowledge embedding [16]. The method of directly extracting governing equations from observational and experimental data using machine learning algorithms to motivate scientific investigation is known as knowledge discovery. The act of incorporating domain knowledge into data-driven models in order to generate models with common sense, increase model accuracy and robustness, reduce data requirements, and create machine learning applications is known as knowledge embedding. Our work falls under the category of knowledge embedding. Moreover, studies that aim to incorporate domain knowledge into machine learning can be generally classified into three categories: data preprocessing, penalty and reward design, and model structure design (see Table 3). Our work falls under both the categories of data preprocessing and model structure design. [Insert Table 3 here]

In IS, knowledge-aware machine learning is starting to gain due attention and has been used to design novel domain-adapted machine learning artifacts. For example, [83] employs psycholinguistics theories to construct a framework that combines domain-adapted NLP artifacts with deep learning models to predict individuals' personalities. In healthcare, the heterogeneity of patient cohorts, the complexity of medical knowledge, and the high need for interpretability contribute to the complexity of healthcare-related learning tasks, where knowledge-aware machine learning can have enormous potential [32].

This study aims to develop a new design artifact for depression detection by utilizing existing medical knowledge on depression. The proposed model incorporates medical knowledge in the form of (1) known symptoms of depressed patients, the life events that cause or exacerbate depression, and medication use; (2) the temporal information and frequency of depression diagnosis-related entities. Our model is expected to effectively identify social media users that have risks of depression. Meanwhile, the identified diagnosis-related entities can explain why a social media user is classified as having risks of depression by highlighting specific depressive symptoms and detected risk factors. This, in turn, can facilitate early intervention measures.

### 2.4. Related Machine Learning and NLP Methods

#### 2.4.1. Depression Ontology (R1)

Ontology is the science of what is, including the types and structures of objects, properties, events, processes, and relationships [67]. Ontology is widely used in computer and information science to provide a standard vocabulary for researchers that need to share information. It provides machine-interpretable definitions of fundamental concepts of the domain and relations between the concepts [67]. In medical expert systems, ontology has been used to represent medical domain knowledge for disease diagnosis [7,85]. Additionally, uncertainty is widely acknowledged in medical expert systems. The Bayesian network and ontology have been integrated to address clinical practice uncertainty and enhance decision-making [85]. For depression detection, ontology-based approaches have been used to represent depression diagnosis terminologies [12,34].

To address our first research question – leveraging medical domain knowledge for depression detection – we construct a depression ontology model that explicitly explains the terminologies used in depression diagnosis and treatments. We also incorporate the prevalence of these terminologies among depression patients into the ontology using the Bayesian network.

**2.4.2. Depression Diagnosis-related Entity Identification (R2)**

Named entity recognition (NER) is a well-suited approach to address the second research question – extracting depression diagnosis-related entities including depression symptoms, major life events, and depression treatments. NER is the task of identifying entities such as people, location, organization, drugs, medical notions, etc [52], which is applied in question answering, information retrieval, coreference resolution, and topic modeling, among others. Hand-crafted rules, lexicons, orthographic features, and ontologies are used in early NER systems followed by feature engineering and machine learning techniques [52]. Later, deep learning-based NER systems with minimal feature engineering have grown in popularity. Such deep learning-based models are useful because they often do not require domain-specific resources (e.g., lexicons), making them more domain-independent [82].

In this study, we adapt the state-of-the-art NER algorithm to identify depression diagnosis-related entities in social media posts. According to the medical literature, the clinical diagnosis of depression is normally based on the chief complaint presented by depressed patients. In the social media context, the following three aspects are often self-reported by social media users and thus detectable: (1) symptoms, such as anxiety, fatigue, low mood, reduced self-esteem, change in appetite or sleep, suicide attempt, etc. [5,9,50,64]; (2) major life event changes, such as divorce, body shape, violence, abuse, drug or alcohol use, and so on [9]; and (3) treatments, the mainstay of depression treatment is usually medication, therapy, or a combination of these two [9]. We, therefore, leverage NER to discover depression symptoms, major life events that may cause or exacerbate depression, and depression treatments.

### 2.4.3. Knowledge-aware Sequence Model for Depression Detection (R3)

Depression evaluation in the medical setting often relies on self-reported depression-related symptoms, major changes in life events, and treatments. In line with this, the objective of this study is to design a model structure that can effectively simulate the superimposed symptoms and signs of depression, as well as capture different progression patterns among social media users by recording the temporal information of depression diagnosis-related entities. Recurrent neural networks (RNN) are well-suited for this aim. An RNN is a deep learning method in which the hidden units are connected in a directed cycle, allowing the network to store past hidden states of information in the internal memory [79]. Long Short-Term Memory (LSTM) is a popular variant of RNN that has a gated structure to handle long-term dependencies [79]. This study uses depression diagnosis-related entities as the inputs for depression detection. While Transformer can be an alternative to the base sequence model, the LSTM-based model empirically outperforms the Transformer-based model, which we will show in the empirical analysis. Therefore, we build the sequence model upon LSTM.

The recency and frequency of the depression diagnosis-related entities are important for depression detection. First, the occurrences of recent major life event changes are reported by the majority of patients with severe depression [70]. Second, LSTM assumes that there is a consistent consecutive property among the input elements, which does not hold in depression diagnosis-related entities for two reasons. (1) The frequency and the number of depression diagnosis-related entities that can be identified in social media data are variable and unstructured because of the irregularity of depressive episodes. (2) Missing information is common in longitudinal social media data because social media users may not necessarily report their depression diagnosis-related symptoms, life events, and treatments.

The relevancy of the depression diagnosis-related entities is also vital in detecting depression.

According to medical domain knowledge [5,9,50,64], different entities carry varying relevance to a depression diagnosis. For example, the entities we can identify on social media may be negative sentiments that do not necessarily indicate depression, because everyone can feel depressed, sad, or blue at some point in their lives. Entities like recurrent thoughts of death and excessive or inappropriate guilt, on the other hand, are strong indicators of depression. Depression diagnosis-related entities may also have varying effects on the onset and progression of depression. For example, traumatic events or major life changes can often trigger depression. Such knowledge (i.e., the relevancy of entities to depression) can be represented in the Bayesian network-based depression ontology. Nonetheless, the terminologies used in the depression ontology (i.e., medical terms) differ significantly from the entities identified on social media (i.e., informal language). We further apply the concept normalization technique to match the entities from social media and diagnosis terminologies in depression ontology [24].

Based on the assumption that more recent and relevant entities are more important in depression detection, we propose the knowledge-aware sequence method to address our third research question – characterizing the temporal and relevance information of depression diagnosis-related entities for depression detection.

## 3. Research Design

To incorporate medical domain knowledge, we first build a depression ontology model that explicitly explains the terminologies (such as various symptoms, life events, and treatments) used in depression diagnosis [19,53]. We also add the prevalence of these terminologies among patients diagnosed with depression into the ontology. We then identify the entities directly associated with depression diagnoses in social media posts, including depression symptoms, life events that may cause or exacerbate depression, and depression treatments [5,9,50,64]. Next, we derive knowledge associated with the identified entities: the temporal and relevancy information. The *temporal information* is the recency and frequency of the entities. The *relevancy information* is from medical knowledge and a pre-trained language model reflecting the semantic textual similarity between the entities and terminologies in depression ontology. Then, we use the entities as the input to perform a knowledge-aware deep learning model for depression detection. Lastly, we perform the attention mechanism on the entities' *temporal* and *relevancy information*. The results can guide interventions for social media users who are at risk of clinical depression and, eventually, have the potential to reduce the societal burden of depression. Figure 1 outlines the flow of our research framework. We articulate the details of each component in the following subsections. [Insert Figure 1 here]

**3.1. Problem Formulation**

In a given social media platform, we collect data from a user base $U$. For a given period of time, we observe the digital traces of the focal user $u \in U$ from $N$ social media posts, denoted by $\left(p_1, p_2, ..., p_N\right)$, ordered in time. Our objectives are two-folded: (1) As summarized in the literature review, depression diagnosis-related symptoms, life events, and treatments are critical indicators of depression and are essential for timely intervention [5,9,50,64]. We aim to detect all the depression diagnosis-related entities from the social media posts, denoted by $\left(le_1, le_2, ..., le_M\right)$, ordered in time. Each entity is the exact phrase as used in the social media posts. (2) According to medical domain knowledge, different depression diagnosis-related entities possess distinct impacts or relevance in depression, we aim to develop a binary classifier with medical domain knowledge that maps $\left(le_1, le_2, ..., le_M\right)$ into two classes: depressed and non-depressed. To achieve these two objectives, we propose the Deep Knowledge-aware Depression Detection Framework (referred to as DKDD hereafter) to detect depression from digital traces on social media. The important notations are listed in Table 4. [Insert Table 4 here]

DKDD features three modules: depression diagnosis-related entities extraction, depression ontology construction, and depression detection. The first module extracts the pre-defined entities that are informative of depression based on medical knowledge. The second module establishes the medical knowledge foundation for our detection and provides the source of depression knowledge for the third module. The third module leverages the knowledge from the first two modules and develops a knowledge-aware attention-based sequence model for depression detection. The specific model components in these modules are shown in the first column in Table 5. The design rationale, the related medical knowledge, and the research gaps addressed for each model component are articulated in the second and third columns in Table 5. Figure 2 shows the architecture of the proposed DKDD framework. [Insert Table 5 & Figure 2 here]

**3.2. Module 1: Depression Diagnosis-related Entities Extraction**

Corresponding to the upper left block of Figure 2, we define depression diagnosis-related entities as personal encounters related to symptoms, major life events, and depression treatments [5,9,50,64]. Figure 3 shows an example. To extract these entities from social media posts, we leverage a state-of-the-art transition-based NER model that has an architecture that chunks and labels a sequence of inputs using Stack-LSTM and allows the NER model to work like a stack that maintains a "summary embedding" of its input [82]. The text representation scheme of this model is RoBERTa [45], as this is the upgraded version of the commonly adopted representation model BERT and empirically performs better than other representation models in our problem.

Let $\left(p_1, p_2, \ldots, p_N\right)$ denote the social media posts of an individual. The resulting model is the depression diagnosis-related entities as phrases $\left(le_1, le_2, \ldots, le_M\right)$. The representation for these depression diagnosis-related entities is $\left(\tilde{x}_1, \tilde{x}_2, \ldots, \tilde{x}_M\right)$, where

$$\tilde{x}_i = RoBERTa(le_i) \tag{1}$$

When describing depression diagnosis-related entities in social media, users may report experiences of families and friends, which are unrelated to their own depression status. To address this issue, we compute an indicator for each entity that denotes whether an entity is experienced by the same user using the following rules: (1) Locate the sentence where entity $le_i$ appears and utilize the SpaCy dependency tree parser[3] to generate the parser tree. (2) Identify the verb of $le_i$. If $le_i$ is a clause, then identify the verb of the corresponding clause. (3) Identify the subject of the verb in step 2 and check whether it is a first-person noun. The final representation for $le_i$ is:

$$x_i = \left(\widetilde{x_i}, firstperson\right),\ firstperson \in \{0, 1\} \tag{2}$$

Unlike prior studies that use the raw social media post as the input, we leverage the distilled depression diagnosis-related entities as our input. These entities are more clinically relevant and practically meaningful for an explanation. Understanding what depression diagnosis-related entities are important in detecting depression is also critical for interventions. To name a few, if entities related to family emergencies are found to be salient predictors of depression detection, the social media platform can act on this information by offering online solutions for emergencies and resources for support groups. If entities related to adverse drug events greatly contribute to the detection, the platform can recommend guidelines from trusted health organizations (e.g., CDC, NIH, and WHO) on what actions to take to alleviate such adverse events. [Insert Figure 3 here]

**3.3. Module 2: Depression Ontology Construction**

The depression diagnosis-related entities extracted in module 1 have varying contributions to predicting a social media user's depression status. Therefore, we develop a depression diagnosis-related ontology to extract and organize entities associated with depression diagnosis and their respective contributions to the diagnostic process from existing medical knowledge.

The motivation behind this step is as follows: if a depression-diagnosis-related entity ($x_i$, extracted from social media posts) rarely occurs among real depression patients, its relevance in predicting depression is low. The ontology we aim to develop focuses on specific aspects of depression, particularly the medical terminologies used in diagnosing depression that are possible to detect from social media posts, including Symptoms, Life events, and Treatment [5,9,50,64], formally denoted as $O_k \in \{Symptom,\ Life\ event,\ Treatment\}$. The purpose of the ontology is to facilitate the detection of depressive symptoms, major life events, and treatments from social media posts. Based on the literature review [4,8,53,67], we compile a list of concepts, $o_j$, related to depression diagnosis (e.g., dejected mood, self-blame, fatal illness, psychotherapy, etc.). Next, we organize the terminologies $o_j$ into three classes: Symptom ($O_{Symptom}$, a collection of depression symptoms), Life event ($O_{Life\ event}$, a collection of major life event changes that may cause or exacerbate depression), or Treatment ($O_{Treatment}$, antidepressants and depression therapies). Meanwhile, we determine the relationships between terminologies ($o_j$) and classes ($O_k$) as $o_j$ "$is\ a$" $O_k$ (e.g., $o_{Dejected\ mood}$ "$is\ a$" $O_{Symptoms}$).

[3] https://spacy.io/api/dependencyparser

In general, the standard ontological model is unable to adequately express domain-specific uncertainty. To address this limitation and concretize the relevancy of terminologies ($o_j$) in depression detection from medical domain knowledge, we build upon previous research [11] and adopt the OntoBayes model, which integrates the Bayesian approach into ontologies. We extend a frequency attribute, $f$, into the ontology. Formally, we denote $o_j$ "$is\ a$" $O_k$, $freq$: $f$, the frequency $f$ is a representation of the population-level occurrence as indicated in the literature among clinically diagnosed depression patients who experience $o_j$. Next, based on the created OntoBayes model, we can deduce that if a patient exhibits $o_j$, the relevance of $o_j$ to depression detection is $f$. For example, $Dejected\ mood$ "$is\ a$" $Symptom, freq$: $0.90$ shows, in medical literature, among clinical diagnosed depression patients, 90% of patients have a "Dejected mood". In turn, in our depression prediction process, if one has a "Dejected mood," it means this person may have a depression "Symptom" and hence suffer from "Depression." The relevancy of "Dejected mood" to depression prediction is 0.90. The evaluation of the depression ontology is presented in the Appendix.

### 3.4. Module 3: Knowledge-aware Depression Detection

We utilize the depression diagnosis-related entities sequence $\left(x_1, x_2, ..., x_M\right)$ obtained from module 1 to predict the depression status of the focal user. This entity sequence encodes more relevant and practically more meaningful depression information than the original social media posts. The conventional LSTM model treats each depression diagnosis-related entity equally. Therefore, each entity would contribute equal information to the detection. In reality, the importance of each entity based on the depression knowledge (i.e., relevancy to depression diagnosis and its time recency) is essential knowledge for depression detection. Based on this observation, DKDD embeds such depression knowledge into our sequence model. For each depression diagnosis-related entity $x_i$, we first learn an effective representation $h_i$ via an LSTM encoder. $h_*$ denotes the user-level representation of the focal user computed as the last hidden unit of the LSTM sequence encoder. We choose LSTM as the encoder because it empirically performs better than other state-of-the-art sequence models, which we will show in the empirical analysis.

$$h_i = f^{(lstm)}\left(h_{i-1}, x_i\right) \quad (3)$$

$$h_* = f^{(lstm)}\left(x_1, x_2, ..., x_M\right) \quad (4)$$

For each encoded information $h_i$ from $x_i$, we aim to weigh it via a depression knowledge-aware attention mechanism. This attention mechanism carries two types of attention weights: temporal attention and ontology attention. We first design the temporal attention. Let $t_T$ be the decision time, and $t_i$ be the time of a depression diagnosis-related entity. The time recency of entity $x_i$ is:

$$\tau_i = t_T - t_i,\ \tau_i \in \Gamma \quad (5)$$

This time recency $\tau_i$ is a scalar. To accommodate the attention computation, we aim to vectorize $\tau_i$ and devise a temporal embedding for $x_i$. Studies have shown that any temporal information can be decomposed into a summation of sine and cosine curves in the frequency domain [73].

Following previous research, we compute the vectorized embedding of $\tau_i$ as:

$$\Phi(\tau_i) = \sqrt{\frac{1}{n_d}}\left[cos\left(w_1\tau_i + \theta_1\right), cos\left(w_2\tau_i + \theta_2\right), \ldots, cos\left(w_{n_d}\tau_i + \theta_{n_d}\right), sin\left(w_1\tau_i + \theta_1\right), sin\left(w_2\tau_i + \theta_2\right), \ldots, si \right. \quad (6)$$

This temporal embedding is useful for distinguishing the importance of entities occurring at different time points. For instance, the attention mechanism may learn to let recent depression diagnosis-related entities stand out because they have a more salient impact on the depressive state. To encode the time recency information of entity $x_i$ into the model, we concatenate the temporal embedding and the hidden representation of $x_i$, which produces self-attention over time, as shown in Equations 7 and 8. Entities with high temporal attention scores have the most salient time recency effect on the depression status at decision time $t_T$.

$$\alpha_i^{(temp)} = tanh\left(x_i \cdot \left[h_i, \Phi\left(\tau_i\right)\right]\right) \quad (7)$$

$$\beta_i^{(temp)} = softmax\left(\alpha_i^{(temp)}\right) = \frac{exp\left(\alpha_i^{(temp)}\right)}{\sum_j^T exp\left(\alpha_j^{(temp)}\right)} \quad (8)$$

Then, we design the ontology attention. According to prior knowledge in the depression ontology constructed in module 2, every depression diagnosis-related entity has varied relevance to depression at the population level. To measure the depression relevance of $x_i$ based on the depression ontology, we identify $x_i$'s most similar entity (e.g., symptoms, life events, and treatments) as indexed in the ontology using vector space-based concept normalization technique [24]. Specifically, we compute the semantic similarity (i.e., cosine similarity) between $x_i$ (i.e., free-text social media language) and all ontology terminologies $o_j \in O$ related to depression (i.e., medical terminologies):

$$sim\left(x_i, o_j\right) = \frac{x_i \cdot o_j}{\left\|x_i\right\|\left\|o_j\right\|} \quad (9)$$

The production of the relevance of $o_j$ to depression ontology $j$ ($freq\left(o_j\right)$) and the semantic similarity between $x_i$ and $o_j$ (i.e., $sim\left(x_i, o_j\right)$) can be used to approximate the relevance of $x_i$ to depression ontology $j$, denoted as $\alpha_i^j$, shown in Equation 10.

$$\alpha_i^j = sim(x_i, o_j) \cdot freq(o_j) \tag{10}$$

Take a real-world scenario as an example: if we identify a depression diagnosis-related entity on social media such as “sad,” the corresponding medical term for similar symptoms in depression ontology would be “dejected mood.” By calculating their semantic similarity using the pre-trained language model (i.e., BERT text representation), we find that “sad” and “dejected mood” are highly related with a similarity score of 0.75 (Equations 9). At the population level, approximately 90% (from Module 2’ OntoBayes model) of depression patients exhibit the symptom of “dejected mood.” Based on such information, the relevance of the depression diagnosis-related entity “sad” to depression classification in our study can be calculated as 0.75×0.9=0.675 (Equations 10).

To encode the medical ontology knowledge into the model, we feed all the resulting relevancy $\left(\alpha_i^1, \alpha_i^2, \ldots, \alpha_i^j, \ldots\right)$ to a neural network, which produces a user-level ontology weight $\alpha_i^{(ont)}$. This weight is further passed to a softmax activation to compute a self-attention score $\beta_i^{(ont)}$, shown in Equations 11-12. Entities with high ontology attention scores have the closest relevance to depression, according to prior knowledge at the population level.

$$\alpha_i^{(ont)} = f^{(mlp)}\left(\alpha_i^1, \alpha_i^2, \ldots, \alpha_i^j, \ldots\right),\ j \in O \tag{11}$$

$$\beta_i^{(ont)} = softmax\left(\alpha_i^{(ont)}\right) = \frac{exp\left(\alpha_i^{(ont)}\right)}{\sum_j^T exp\left(\alpha_j^{(ont)}\right)} \tag{12}$$

In the end, we fuse the temporal attention and the ontology attention. Such temporal knowledge and ontology knowledge are distinct for various users. Therefore, it is essential to consider the user-level heterogeneity when fusing this knowledge. Recall that Equation 4 produces a user-level representation of the focal user, we aim to leverage this representation to perform an attention mechanism to assign varied temporal-ontology weights to different users.

We feed the user-level representation $h_*$ to a neural network, which results in a two-dimensional vector $\left(a_1, a_2\right)$. This vector is user-specific. We apply a softmax activation to this vector to make sure its two dimensions sum up to one. Such normalized two dimensions $\left(\widetilde{a}_1, \widetilde{a}_2\right)$ can serve as the fusion weights for the temporal and ontology attention. Thus, the temporal and ontology knowledge are fused via a convex combination of two weights to attend depression-diagnosis-related entities, as shown in Equation 15. This allows each entity to be embodied with two critical types of depression knowledge, instead of learning to make a classification from scratch each time. We aggregate all entities' hidden representations on the basis of these attention weights. An MLP layer is stacked in the end to predict the depression risk, as shown in Equation 16.

$$\left(a_1, a_2\right) = f^{(mlp)}\left(h_*\right) \tag{13}$$

$$\left(\widetilde{a}_1, \widetilde{a}_2\right) = softmax\left(a_1, a_2\right) \tag{14}$$

$$\widetilde{h} = \sum_{j=1}^{T}\left(\widetilde{a}_1 \cdot \beta_i^{(temp)} + \widetilde{a}_2 \cdot \beta_i^{(ont)}\right) \tag{15}$$

$$y = f^{(mlp)}\left(\widetilde{h}\right) \tag{16}$$

**3.5. The Novelty of the Proposed Framework**

The proposed DKDD framework has the following novelties: (1) Unlike the common approach of using raw social media posts as the input, we distill depression diagnosis-related entities as the input. These entities not only assist model prediction but also provide an explanation of the user's medical history and the potential explanatory factors of depression, which are critical for designing targeted intervention strategies. These entities together with our attention weights and fusion weights serve as a natural interpretation mechanism for our prediction model. (2) We design and incorporate a medical ontology model to enrich medical knowledge in depression detection. Therefore, the relevancy of the depression-related entities that we identified can be better assessed. (3) We develop a comprehensive framework for fusing both temporal knowledge and ontology knowledge in depression detection.

**4. Design Evaluation**

### 4.1. Datasets

Our depression detection testbed comes from the eRisk 2018 (task 1) and eRisk 2020 (task 1) datasets [46]. This dataset contains 1,349,842 archival Reddit posts (29,521,866 words) from 2,470 labeled subjects, including 359 depression patients and 2,111 non-depression users. A summary description of the dataset is shown in Table 6. To train a NER model to detect depression diagnosis-related entities, we select a cross-platform dataset for annotation, including a random sample of 2,500 posts from the eRisk dataset (social media posts from Reddit) and 2,500 posts from WebMD. The NER annotation dataset expands from the depression detection dataset because this ensures the NER annotation is robust in various platforms and can be adopted by other studies. The eRisk dataset is articulated in section 4.1.2. WebMD is a leading health IT platform that attracts a large number of patients to write drug reviews and share their medication-taking experiences. We collected all the posts related to antidepressants because these patients are diagnosed with depression and share the most relevant experience around depression. This cross-platform NER annotation dataset ensures our model can be generalized to monitor digital traces in many health-related social media platforms. [Insert Table 6 here]

### 4.2. Depression Diagnosis-related Entity Extraction Results

As described in Section 4.1., the annotation dataset for the depression diagnosis-related entities extraction model includes 2,500 posts from eRisk and 2,500 posts from WebMD. Four well-trained annotators with a health analytics background manually read these posts and annotated the depression diagnosis-related entities, including disease symptoms, major life events, and depression treatments. Figure 3 shows an example of the annotation. The highlighted phrases are annotated as depression diagnosis-related entities. Another independent expert annotator read through all the annotation data. The Kappa values for the inter-annotator reliability are 0.775 on the eRisk dataset and 0.905 on the WebMed dataset. Table 7 presents the performance evaluation of depression diagnosis-related entity extraction. This model reaches an F1-score of 84.8% in the hold-out test set, indicating excellent performance. We, then, apply this model to our social media posts and extract depression diagnosis-related entities. [Insert Table 7 here]

### 4.3. Depression Detection Results

Each of the following results is the average performance of 20 experiment runs. We also report the standard deviation of these experiments. We use 60% of the data for training, 20% for validation, and 20% for testing. We use the extracted depression diagnosis-related entities in the posts as the input to predict depression status. We compare DKDD with the state-of-the-art depression detection models [11,13,17,20,38,43,49,61,62]. The evaluation results are reported in Tables 8-9.

#### 4.3.1. Comparing DKDD to Benchmark Methods

Table 8 shows the comparison with studies using traditional machine learning methods, and Table 9 reports the comparison with studies using deep learning. For both groups of baseline models, our proposed DKDD shows significant performance improvement (Figure 4 and Figure 5). Compared with the best-performing model in traditional machine learning [62], DKDD improves the AUC score by 0.095. Compared with the best-performing deep learning model [39], DKDD improves the AUC score by 0.059. These baseline models reach relatively low performance mainly due to three reasons: (1) The depression detection task is at the user level instead of the post level. User-level processing is technically more complex which involves analyzing a large number of digital traces of the focal user. However, extracting valid information from these traces is a challenge for traditional machine learning models that rely on feature engineering. On the other hand, standard deep learning models neglect medical knowledge, thus leading to inferior performance to our method. (2) Table 9 also indicates that the LSTM-based model outperforms Transformer in depression detection. Transformers have shown remarkable performance in various machine learning tasks. However, recurrent neural networks are still relevant and find use in many practical scenarios. Despite claims that LSTMs and RNNs are dead, a lot of research suggests otherwise. [36] suggest that vanilla Transformers have quadratic complexity concerning the input’s length, making them prohibitively slow for very long sequences. Furthermore, studies have revealed that Transformers (as a type of feed-forward sequence models) fail to generalize in many simple tasks that recurrent models (such as LSTM) handle with ease when the inputs' length exceeds those observed at training time [22]. As user-level depression detection requires analyzing lengthy historical social media posts, we opt for LSTM as the sequence encoder in Module 3 of DKDD. (3) Our dataset is highly imbalanced, with an imbalance ratio of 1:6. This data is representative of real-world statistics, where the prevalence of adults with a major depressive episode was 17.0% among individuals aged 18-25, a demographic that includes a significant number of social media users. It is very challenging for these baseline models to excel in this context. [Insert Tables 8-9 & Figures 4-5 here]

**4.3.2. Ablation Analysis of DKDD**

The DKDD framework incorporates three types of knowledge: temporal knowledge, ontology knowledge, and depression diagnosis entity knowledge. We remove each type of knowledge from DKDD to test their influence. Removing the diagnosis entity knowledge implies using the raw posts as the input, which is the same as the existing deep learning-based studies in this area. As reported in Table 11, removing any type of knowledge will significantly hamper the detection performance. Specifically, removing the temporal knowledge lowers the F1 score by 0.010, removing the ontology knowledge drops the F1 score by 0.015, and removing the diagnosis entity knowledge reduces the F1 score by 0.039. Such knowledge-induced performance improvement consistently appears in AUC, precision, and recall as well (Figure 6). Among these three types of knowledge, the diagnosis entity knowledge is the most influential, thus supporting our model design. This is our new design as compared to the prior depression detection studies using social media data. These entities are directly related to users' depressive experiences, and noises in raw social media posts are filtered out. [Insert Table 11 & Figure 6 here]

For the above knowledge, we further examine their robustness in different models. Table 12 shows the robustness of temporal knowledge. We report two models here: DKDD and the temporal model. The temporal model removes ontology knowledge from DKDD. The results show that in both of these models, removing the temporal knowledge would decrease the performance in all four evaluation metrics. For DKDD, the temporal knowledge contributes to an F1 gain of 0.010. For the temporal model, the temporal knowledge contributes an even larger performance gain, with an F1 increase of 0.059. The temporal knowledge represents the information about the time-sensitivity of each diagnosis-related entity. Incorporating such knowledge in the model is critical for any time series analysis, such as the depression history in our context. [Insert Table 12 here]

Table 13 shows the robustness of the ontology knowledge. We test its effect on two models: DKDD and the ontology model, where the ontology model is to remove temporal knowledge from DKDD. For both models, removing the ontology knowledge decreases the detection performance in all metrics. The ontology knowledge is attributed to an F1 improvement of 0.016 for DKDD, whereas it improves the F1 of the ontology model by a larger margin (0.064). The ontology knowledge is medical knowledge-driven information. Leveraging this medical knowledge allows the model to learn on top of what is already known in the medical field, without having to learn from scratch or randomly. This ontology-driven design is not only useful in depression detection, but it can assist many other health-related prediction tasks as well, such as disease prediction, readmission prediction, and adverse drug event identification. [Insert Table 13 here]

Table 14 depicts the robustness of the diagnosis entity knowledge. We show its effectiveness in DKDD, the temporal model, and the ontology model. The results indicate that removing the diagnosis entity knowledge from any model will significantly lower their performance in all four metrics. Specifically, the diagnosis entity knowledge increases the F1 for DKDD by 0.039, improves the F1 for the temporal model by 0.048, and raises the F1 for the ontology model by 0.029. These diagnosis entities directly depict the personal depression-related experiences of the users without the noisy information in the original raw social media posts. This approach is in contrast to many end-to-end sequence models that directly take a social media post as the input. Our entity-based approach serves as a filter mechanism and promotes interpretability for the model. It can be fruitfully adapted by other text mining research in IS where historical information needs to be considered. [Insert Table 14 here]

As the diagnosis-related entities are extracted from historical posts, we test the number of entities that are needed for depression detection. We refer to the number of entities as the memory size. As shown in Table 15, when the memory size increases, the model performance improves. After the memory size of 200, increasing the memory size decreases the performance. This is because, when the memory size is small, increasing the size incorporates richer depression history about the user. When the memory size is too large, the far past experiences are outdated. They do not have a material impact on the user's current mental status anymore. Tracing back to those entities would include noise to the model. Therefore, we use the memory size of 200 for all other analyses in this study. [Insert Table 15 here]

**4.3.3. Explainable Insights of Depression Detection**

Since we design an attention-based knowledge-aware (*relevance* and *recency* of depression diagnosis-related entities) model, the attention scores can explain the major contributors to depression detection. We randomly select four subjects in Figure 7 that our model predicted as depressed. The red words are the selected depression diagnosis-related entities that have the highest attention scores in each time interval. Bigger font with darker color suggests the entity has a higher attention score, indicating it is more important in predicting depression. We organize them by the time of the entity, streamlining the series of events that the subject experienced.

The first subject reported digital traces such as symptoms and major life events. This subject experienced weight loss, nausea, and an eating disorder. The negative effects and social stigma around them could induce depression. This subject also reported losing someone in the family, which is a significant contributor to severe depression. The second subject's digital traces include symptoms, major life events, and treatments. This subject was stressed about his or her LGBTQ identity, suffered from ADHD, and had multiple episodes of self-harm attempts. The third subject described digital traces of symptoms and treatments. This subject is on the Autism spectrum and has bipolar disorder. These are chronic diseases that cause long-haul symptoms and require regular treatments. These events pose a significant threat to the subject's mental health. The fourth subject showed digital traces of major life events. This subject experienced abusive behavior from a relationship and had self-esteem issues. Because of these life events, the subject reported multiple attempts of suicide and hurting animals. [Insert Figure 7 here]

The depression diagnosis-related entities discovered in all these four subjects attributed to their depression. Our model fruitfully leveraged them to predict their depression and is able to explain the most important contributing entities. This has significant practical implications as social media platforms can identify those at risk of depression in a timely manner and understand the salient contributing factors. Corresponding intervention recommendations can be sent to the users to address those factors.

## 5. Discussion and Future Research

Depression is a critical global issue. It is one of the leading causes of disability and death, resulting in substantial social and economic burdens, including decreased productivity and increased healthcare expenses. Depression detection and interventions are crucial to address this societal and health concern. With the advent of digital trace data, researchers can identify patterns and trends that are indicative of various disorders, including depression, and predict their occurrences based on associated behavioral or psychological changes, signs, and symptoms. Following the design science research paradigm [27,31], this study identifies entities in social media posts that are directly related to depression diagnosis and develops a knowledge-aware deep learning framework to accurately detect social media users with or at risk of depression. Empirical evaluations using real-world datasets demonstrate that our proposed DKDD outperforms state-of-the-art techniques. Moreover, DKDD provides meaningful insights into the factors that contribute to the detection of depression in social media users, which can inform the development of effective intervention strategies.

### 5.1. Research Contributions and Implications to IS Research

From the perspective of design science, we make two contributions. First, we propose a novel deep knowledge-aware framework for detecting depression, which leverages user-generated digital trace data. Second, as part of our framework, we propose three novel IT artifacts, which include (1) the extraction of depression diagnosis-related entities to remove noisy information and improve interpretability, (2) the integration of a medical ontology model into the detection model to enhance model performance, and (3) a comprehensive framework for fusing both input data's temporal knowledge and ontology knowledge.

Our method and results also have three implications for IS research. (1) Knowledge-aware machine learning presents numerous opportunities and is a promising area of IS study. Current trends in machine learning involve developing larger models that require extensive amounts of training data, which can be a challenge for individual researchers or those working in small institutions. However, knowledge-aware machine learning provides an alternative by incorporating domain knowledge into the model design to develop more efficient model structures and to provide better interpretability [83,86]. Knowledge-aware machine learning can be valuable in many applications that involve complex learning tasks and limited training resources. Our modeling framework and design principles offer a nascent design theory [27], which can inspire other scholars to explore the potential of this direction. (2) Using digital trace data to address social concerns is a growing field in IS research. As the Fourth Industrial Revolution becomes actual, IS researchers can access vast amounts of digital trace data that offers valuable insights into human behavior and activities. Our work focuses on using social media users' digital traces for depression detection. However, we believe that the idea of using digital trace data to address social concerns is applicable to other problem domains as well. For example, digital trace data can also be used to detect other mental health issues, disease outbreaks, public opinion trends, crime and safety, etc.

Overall, our work demonstrates that digital trace data can provide valuable insights into social concerns, and by combining domain knowledge with specific research context, IS researchers can design innovative "machine learning artifacts in context [56]" to solve business and societal problems. (3) Our work belongs to the NLP research in IS and aligns with the perspective of previous IS design science studies, which emphasize the importance of designing artifacts related to text and social media [2,14,83]. NLP enables researchers to extract valuable information in the form of new attributes from vast amounts of unstructured text data, thereby bridging the intersection of design and data science. By incorporating these attributes into analysis, researchers can gain a deeper understanding of various societal and business problems. As such, NLP holds significant potential for advancing research in IS to address various socio-technical concerns.

### 5.2. Practical Implications

Depression is a serious global health concern that requires attention and action from individuals, online communities, and public health administration. The practical implications of this study are three-fold. (1) First, on the individual level, although, depression is a complex mental health disorder that can have multiple causes, including genetic, environmental, and psychological factors, which may require a multifaceted approach for prevention, our approach allows users to be aware early of their risk for depression and provides insight into the reasons for such a prediction. With this understanding, individuals can proactively develop healthier lifestyles that may improve depressive symptoms and, more importantly, seek help early if they are unable to manage the problem alone. (2) Second, our approach enables social media platforms to monitor user behavior and identify patterns that may indicate a user's risk of depression. By detecting and providing support and resources to at-risk users before they reach a crisis point, social media platforms can help prevent depression. Our model also provides insights into the specific reasons why a user may be at risk of depression, allowing platforms to increase awareness and education about mental health by sharing relevant information. For example, platforms can partner with mental health organizations to promote their services, thereby increasing access to mental health resources. In addition, our framework can identify online harassment and bullying as factors contributing to a user's depression. Platforms can take steps to reduce these behaviors and promote a positive and supportive online environment. This includes decreasing the stigma associated with depression, providing tools to report and block harmful content, promoting positive mental health messages, and creating safe spaces for users to seek support and connect with others who may be experiencing similar challenges. (3) Lastly, given that our method uses readily available digital trace data generated by social media users, it enables large-scale depression detection at the population level. Thus, public administration and policymakers can utilize our framework and detection results in a number of ways. First, our framework can aid in creating evidence-based policies tailored to the specific needs of different groups of people. For instance, resources can be allocated by policymakers toward providing mental health services to populations affected by natural disasters. Secondly, the analysis results can inform public health campaigns aimed at promoting mental health, reducing stigma, bullying, and online harassment, advocating for increased mental health research funding, and raising awareness of the importance of mental health in the public sphere.

### 5.3. Limitations and Future Work

Although promising, this study has several limitations. The first limitation stems from the use of social media data. Although social media is valuable for depression detection, social media users are not an entirely representative sample of the target population. Also, the social media dataset has potential selection bias by relying on self-reported depression patients to share their diagnoses publicly [61]. To address the selection bias of social media, we propose using our method to provide complementary information to existing clinical depression screening procedures. The second limitation lies in the potential ethical and privacy issues. For example, the issuance companies or companies' recruiters may use the prediction results against individuals who suffer from depression or are at risk of depression. Our method can be extended in the following directions. First, we plan to use NLP, computer vision, and other techniques to extract information from text, images, and other forms of digital trace data. The fusion of these multiple modes of data could capture a more complete picture of Internet users at risk of mental disorders. Second, depression has a snowball effect on peers. Digital traces from social media often describe social relationships whereas conventional survey techniques only measure the characteristics of individual subjects. We can include network analysis in our future work to measure the network effects of depression [78].

## 6. Conclusion

As the Fourth Industrial Revolution transitions from a buzzword to reality, more and more digital trace data provides IS researchers valuable insights into social concerns and can be used to develop innovative machine learning artifacts in context to address business and societal problems. Meanwhile, knowledge-aware machine learning is a promising area of IS study that incorporates domain knowledge into the model design to develop efficient model structures with better interpretability, making it valuable in many applications with limited training resources. By proposing a deep knowledge-aware depression detection framework using digital trace data on social media, this study demonstrates the potential of IS research in addressing important societal problems like depression. Our proposed framework innovatively incorporates three categories of medical knowledge: depression diagnosis-related entities, temporal knowledge, and medical ontology knowledge. Extensive empirical studies with real-world data demonstrate that, by considering such knowledge, our framework achieves superior performance compared to numerous existing depression detection works. Leveraging the depression diagnosis-related entities and our attention mechanism, our framework is able to explain the critical factors that contribute to depression detection. This framework is a deployable tool for many stakeholders, including social media platforms, users, and policymakers.

## Tables

**Table 1. Depression-related Social Media Posts**

| Platform | Sample Post | Digital Traces (Cat *) | | |
|---|---|---|---|---|
| | | 1 | 2 | 3 |
| reddit | Posted by 1 hour ago<br>Depressed in Mexican home<br>They don't believe in depression they believe in laziness, selfishness, ungratefulness. If only they knew how I just don't have the energy Im physically tired and exhausted. If I | ✔ | | |
| | Posted by 1 month ago 2<br>(Oc) I've been diagnosed with depression 3 months ago and have really low self esteem so I've decided to collect all the nice comments I've | ✔ | | |
| | Posted by 1 month ago<br>I'm having panick attacks and bouts of depression because of my relationship | ✔ | ✔ | |
| facebook | March 15, 2021 · ...<br>I act happy but i wanna die | ✔ | | |
| | April 18, 2021 · ...<br>Maybe I don't deserve to be happy<br>Maybe I deserve to be hated by everyone | ✔ | | |
| twitter | · Feb 14 ...<br>Had my psych eval today. They're treating me for anxiety and **depression**. Starting out on Buspiron 7.5mg twice a day and **Zoloft** 25mg in the morning.<br>I'm feeling nervous about that whole thing but I think it's a good step. I have to try. | | | ✔ |
| | · Feb 22<br>For anyone who needs to hear this: If you're anxious or depressed, please get help. Go to counseling. Talk about it. I'm only saying this cuz I'm leaving my monthly psychiatrist visit. So many of us are not OK, so let's be not OK together. #MentalHealth | ✔ | | ✔ |

Note: Cat 1. = Symptoms; Cat 2. = Major life events; Cat 3. = Treatments

**Table 2. Recent Social Media-based Mental Health Detection Methods and Identified Research Gaps**

| Reference | Model | Features | Research Gaps and Advantages |
|---|---|---|---|
| [11,13,17, 20,38,43,49,61,62] | Traditional ML methods: SVM, Regression Random Forests, XGBoost, Naïve Bayes, | LIWC, n-grams, sentiment scores, topics, social media metadata. | • Features do not specify clinical depression.<br>• Inadequate performance. |
| [18,21,69] | Rule-based classification, LDA, Neural Network | Social media users' online activities and behavioral patterns. | ● Features used are considerably different from the medical definition of depression.<br>● Inadequate performance. |
| [38,43,49] | Pure Deep Learning models:<br>CNN, LSTM, Transformer | Raw social media posts with representation learning. | ● Limited interpretability: difficult to understand the reasons behind the models' classification of an individual as depressed.<br>● Massive labeled data is required for training. |
| DKDD (ours) | Knowledge- aware deep learning model | Depression diagnosis-related entities. | ● Elevated detection performance.<br>● Explain why a social media user is classified as depressed and facilitate early intervention. |

**Table 3. Representative Knowledge-aware Machine Learning vs. Our Method**

| Method | Domain Knowledge Incorporated | Categories | | |
|---|---|---|---|---|
| | | Data Preprocessing | Penalty and Reward Design | Model Design |
| [58] | Classification theory | | | ✔ |
| [6] | Behavioral theory | | | ✔ |
| [39] | Relationship marketing theory | ✔ | | |
| [74] | Expert experience in engineering control | | ✔ | |
| [41] | Behavioral knowledge in ontology | | | ✔ |
| [32] | Epidemiology knowledge | | | ✔ |
| [83] | Psycholinguistics theories | | | ✔ |
| DKDD (ours) | Medical domain knowledge in clinical depression detection | ✔ | | ✔ |

**Table 4. Important Notations**

| Notation | Description | Notation | Description |
|---|---|---|---|
| $p_i$ | The $i$-th post of the focal user | $\beta_i^{(temp)}$ | Attention weight of temporal knowledge |
| $le_i$ | The $i$-th depression diagnosis-related entity of the focal user | $\beta_i^{(ont)}$ | Attention weight of ontology knowledge |
| $x_i$ | The representation of $le_i$ | $\tilde{a}_1$ | Fusion weight of temporal knowledge |
| $h_i$ | The encoded representation of $x_i$ | $\tilde{a}_2$ | Fusion weight of ontology knowledge |
| $h_*$ | User-level representation | $y$ | Depression label |

**Table 5. Design Guidelines for DKDD**

| Model Components | | Medical Knowledge | Research Gaps |
|---|---|---|---|
| Depression diagnosis-related entities extraction | | Depression diagnosis-related symptoms, life events, and treatments are critical indicators of depression and are essential for timely intervention [5,9,50,64]. | Depression diagnosis-related symptoms, life events, and treatments are often overlooked in depression detection studies. |
| Sequence encoder | | Depression diagnosis-related symptoms, life events, and treatments are sequential data. Sequence encoders, such as LSTM, can model such time-series historical events [79]. | Depression diagnosis-related medical knowledge is under-explored in machine learning models for depression detection. |
| Knowledge-aware components | Time | Time recency is essential knowledge to determine depression. More recent depression diagnosis-related symptoms, life events, and treatments have a more salient influence than the older counterparts [70,80]. | |
| | Ontology | Depression ontology information, including symptoms, life events, and treatment, is indicative of depression risk [5,9,50,64]. | |
| | Attention | Different depression diagnosis-related symptoms, life events, and treatments carry distinct influences on depression risk. They can be weighted to provide a more accurate depression prediction [5,9,50,64,80]. | |

**Table 6. Dataset Summary**

| Statistic | Depressed | Non-depressed |
|---|---|---|
| Num. of posts | 108,840 | 1,241,002 |
| Num. of words | 2,902,193 | 26,619,673 |
| Avg. num. of posts per subject | 303 | 588 |
| Avg. num. of days from first to last post | 621.5 | 625.4 |
| Num. of subjects | 359 | 2,111 |

**Table 7. Depression Diagnosis-related Entities Extraction Results**

| Metric | Train | Validation | Test |
|---|---|---|---|
| F1 | 0.848 | 0.840 | 0.848 |
| Precision | 0.896 | 0.885 | 0.932 |
| Recall | 0.805 | 0.799 | 0.777 |

**Table 8. Comparison with Traditional Machine Learning with Feature Engineering**

| Model | AUC | F1 | Precision | Recall |
|---|---|---|---|---|
| DKDD (Ours) | 0.824 ± 0.014 | 0.828 ± 0.012 | 0.836 ± 0.029 | 0.824 ± 0.014 |
| [17] | 0.569 ± 0.001 | 0.588 ± 0.002 | 0.716 ± 0.001 | 0.569 ± 0.003 |
| [20] | 0.685 ± 0.002 | 0.705 ± 0.001 | 0.735 ± 0.003 | 0.685 ± 0.002 |
| [60] | 0.723 ± 0.003 | 0.760 ± 0.002 | 0.820 ± 0.001 | 0.720 ± 0.001 |
| [11] | 0.716 ± 0.029 | 0.722 ± 0.026 | 0.730 ± 0.022 | 0.716 ± 0.029 |
| [62] | 0.729 ± 0.012 | 0.717 ± 0.011 | 0.708 ± 0.013 | 0.729 ± 0.011 |
| [13] | 0.623 ± 0.006 | 0.573 ± 0.005 | 0.570 ± 0.002 | 0.622 ± 0.005 |

**Table 9. Comparison with Deep Learning with Representation Learning**

| Model | AUC | F1 | Precision | Recall |
|---|---|---|---|---|
| DKDD (Ours) | 0.824 ± 0.014 | 0.828 ± 0.012 | 0.836 ± 0.029 | 0.824 ± 0.014 |
| CNN-based [45] | 0.710 ± 0.005 | 0.711 ± 0.004 | 0.728 ± 0.018 | 0.710 ± 0.005 |
| LSTM-based [39] | 0.765 ± 0.004 | 0.756 ± 0.004 | 0.751 ± 0.008 | 0.765 ± 0.004 |
| Transformer-based [51] | 0.751 ± 0.002 | 0.734 ± 0.005 | 0.724 ± 0.010 | 0.751 ± 0.002 |

**Table 11. Ablation Studies**

| Model | AUC | F1 | Precision | Recall |
|---|---|---|---|---|
| DKDD (Ours) | 0.824 ± 0.014 | 0.828 ± 0.012 | 0.836 ± 0.029 | 0.824 ± 0.014 |
| DKDD removing temporal knowledge | 0.811 ± 0.012 | 0.818 ± 0.008 | 0.827 ± 0.016 | 0.811 ± 0.012 |
| DKDD removing ontology knowledge | 0.792 ± 0.016 | 0.813 ± 0.014 | 0.838 ± 0.018 | 0.792 ± 0.016 |
| DKDD removing diagnosis entity knowledge* | 0.794 ± 0.015 | 0.789 ± 0.013 | 0.789 ± 0.023 | 0.794 ± 0.015 |

*Raw post as input.

**Table 12. Robustness of Temporal Knowledge**

| Model | AUC | F1 | Precision | Recall |
|---|---|---|---|---|
| DKDD (Ours) | 0.824 ± 0.014 | 0.828 ± 0.012 | 0.836 ± 0.029 | 0.824 ± 0.014 |
| DKDD removing temporal knowledge | 0.810 ± 0.012 | 0.818 ± 0.008 | 0.827 ± 0.016 | 0.811 ± 0.012 |
| Temporal model* | 0.792 ± 0.016 | 0.813 ± 0.014 | 0.838 ± 0.018 | 0.792 ± 0.016 |
| Temporal model removing temporal knowledge | 0.751 ± 0.016 | 0.754 ± 0.015 | 0.759 ± 0.020 | 0.751 ± 0.016 |

*Temporal model: remove ontology knowledge from DKDD.

**Table 13. Robustness of Ontology Knowledge**

| Model | AUC | F1 | Precision | Recall |
|---|---|---|---|---|
| DKDD (Ours) | 0.824 ± 0.014 | 0.828 ± 0.012 | 0.836 ± 0.029 | 0.824 ± 0.014 |
| DKDD removing ontology knowledge | 0.792 ± 0.016 | 0.812 ± 0.015 | 0.838 ± 0.018 | 0.792 ± 0.016 |
| Ontology model* | 0.811 ± 0.012 | 0.818 ± 0.008 | 0.827 ± 0.016 | 0.811 ± 0.012 |
| Ontology model removing ontology knowledge | 0.751 ± 0.016 | 0.754 ± 0.015 | 0.759 ± 0.021 | 0.751 ± 0.016 |

*Ontology model: remove temporal knowledge from DKDD.

**Table 14. Robustness of Diagnosis Entity Knowledge**

| Model | AUC | F1 | Precision | Recall |
|---|---|---|---|---|
| DKDD (Ours) | 0.824 ± 0.014 | 0.828 ± 0.012 | 0.836 ± 0.029 | 0.824 ± 0.014 |
| DKDD removing diagnosis entity knowledge | 0.794 ± 0.015 | 0.789 ± 0.012 | 0.788 ± 0.023 | 0.793 ± 0.015 |
| Temporal model | 0.792 ± 0.016 | 0.813 ± 0.014 | 0.838 ± 0.017 | 0.792 ± 0.016 |
| Temporal model removing diagnosis entity knowledge | 0.769 ± 0.019 | 0.765 ± 0.017 | 0.764 ± 0.025 | 0.769 ± 0.018 |
| Ontology model | 0.811 ± 0.012 | 0.818 ± 0.008 | 0.827 ± 0.016 | 0.811 ± 0.012 |
| Ontology model removing diagnosis entity knowledge | 0.786 ± 0.011 | 0.789 ± 0.013 | 0.794 ± 0.022 | 0.786 ± 0.011 |

**Table 15. Effect of Memory Size of Diagnosis Entity**

| Memory Size | AUC | F1 | Precision | Recall |
|---|---|---|---|---|
| 50 | 0.783 ± 0.013 | 0.800 ± 0.011 | 0.824 ± 0.031 | 0.783 ± 0.013 |
| 100 | 0.798 ± 0.016 | 0.815 ± 0.014 | 0.838 ± 0.026 | 0.798 ± 0.016 |
| 200 | 0.824 ± 0.014 | 0.828 ± 0.012 | 0.836 ± 0.029 | 0.824 ± 0.014 |
| 300 | 0.821 ± 0.017 | 0.819 ± 0.014 | 0.818 ± 0.023 | 0.821 ± 0.017 |

# Figures

## Figure 1. Flowchart of DKDD

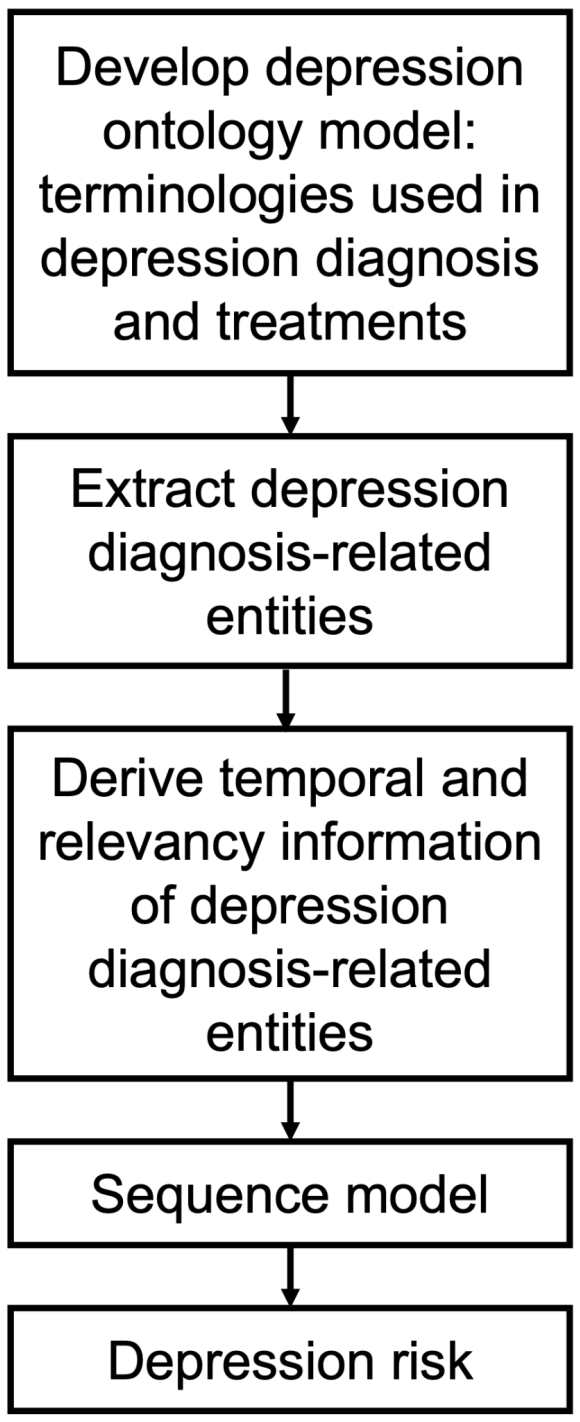

## Figure 2. DKDD Architecture

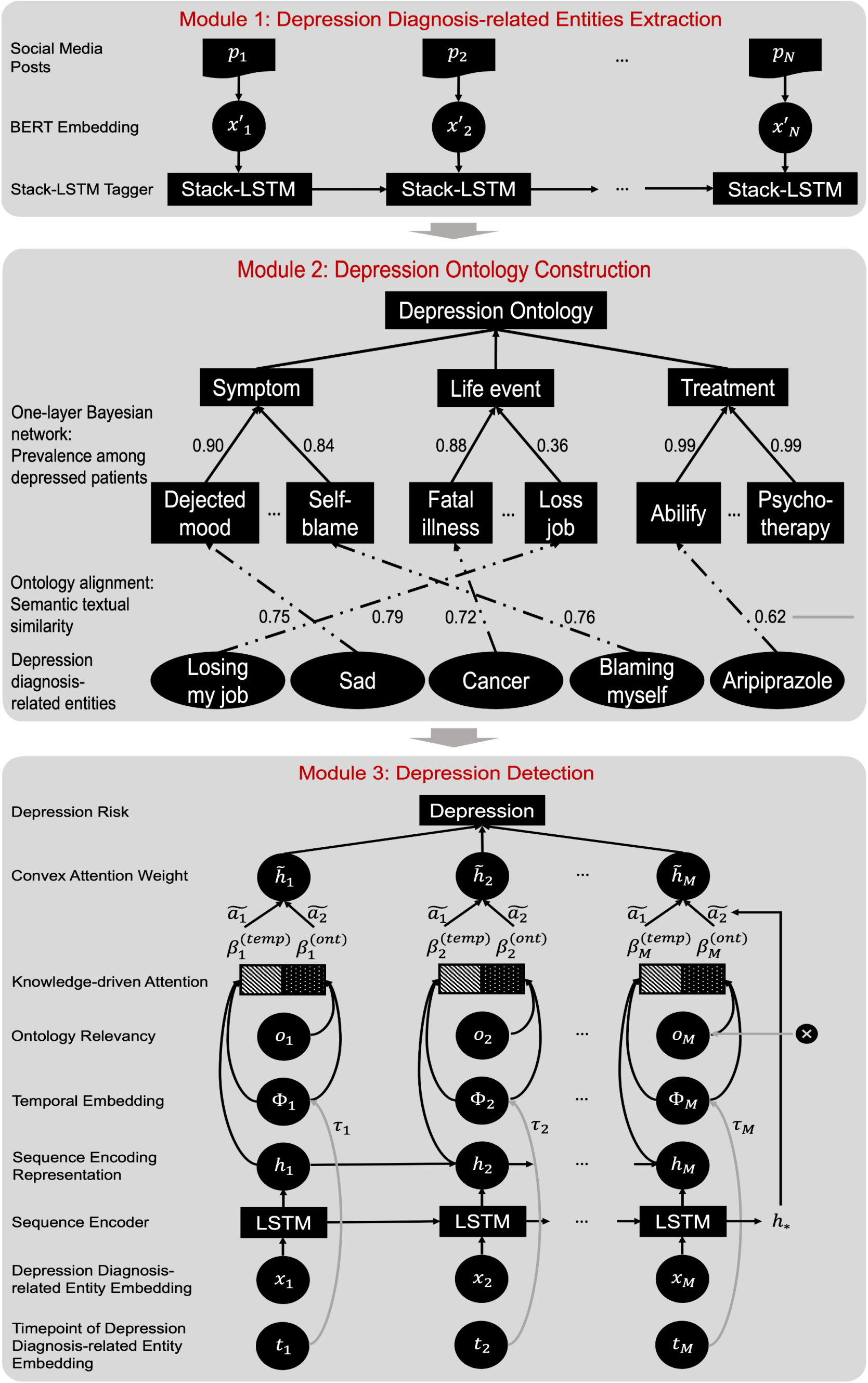

## Figure 3. Depression Diagnosis-related Entities Extraction

This medication Ativant has helped me a lot with anxiety and panic attack. I am also on Serequil and I have to take Ativant to ease the side effects. I become very restless or irritated and the dr. gave me Ativant to try to offset it.

Depression Diagnosis-related Entity

**Figure 4. Improvements over Traditional ML**

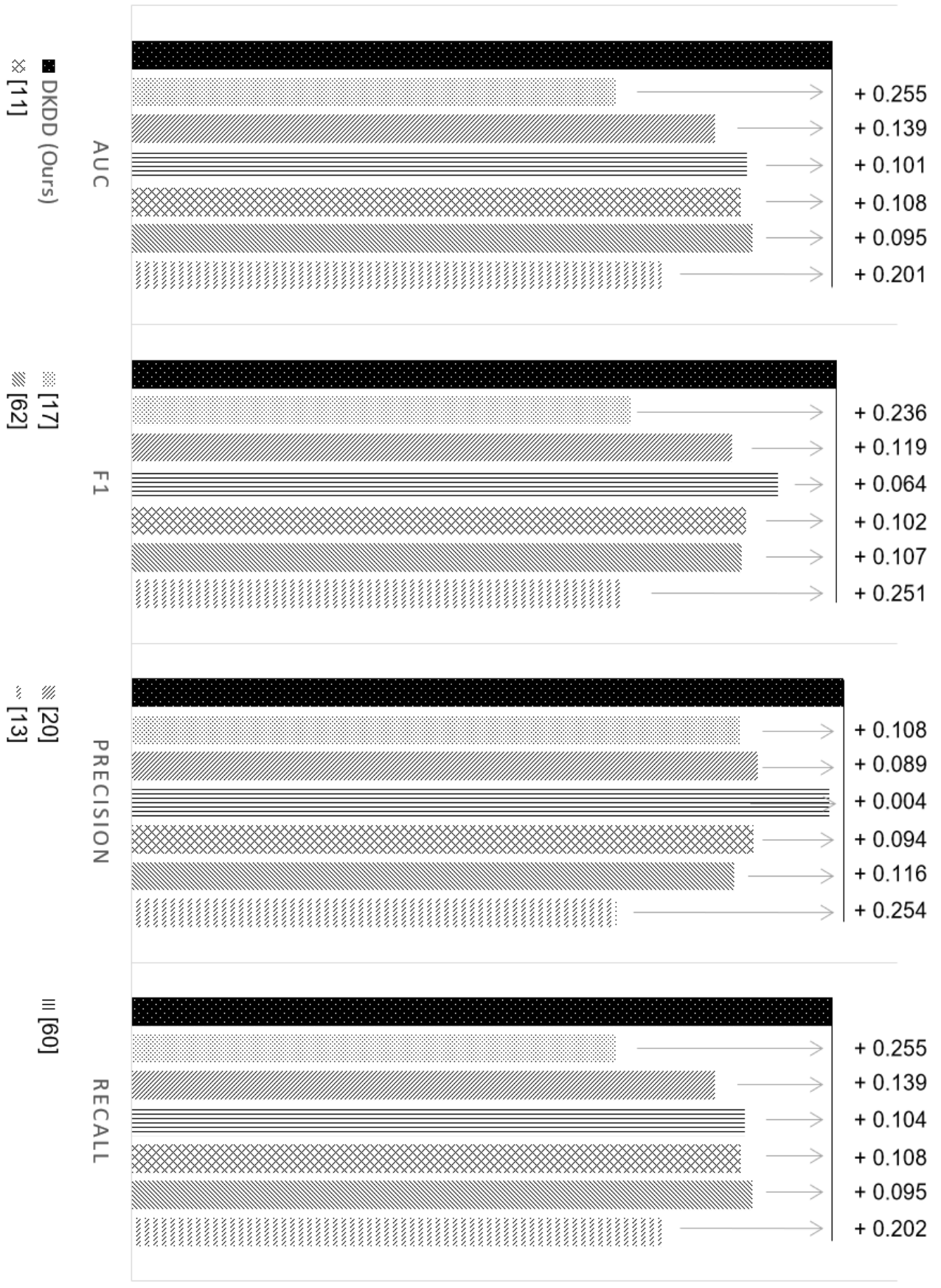

**Figure 5. Improvements over Deep Learning**

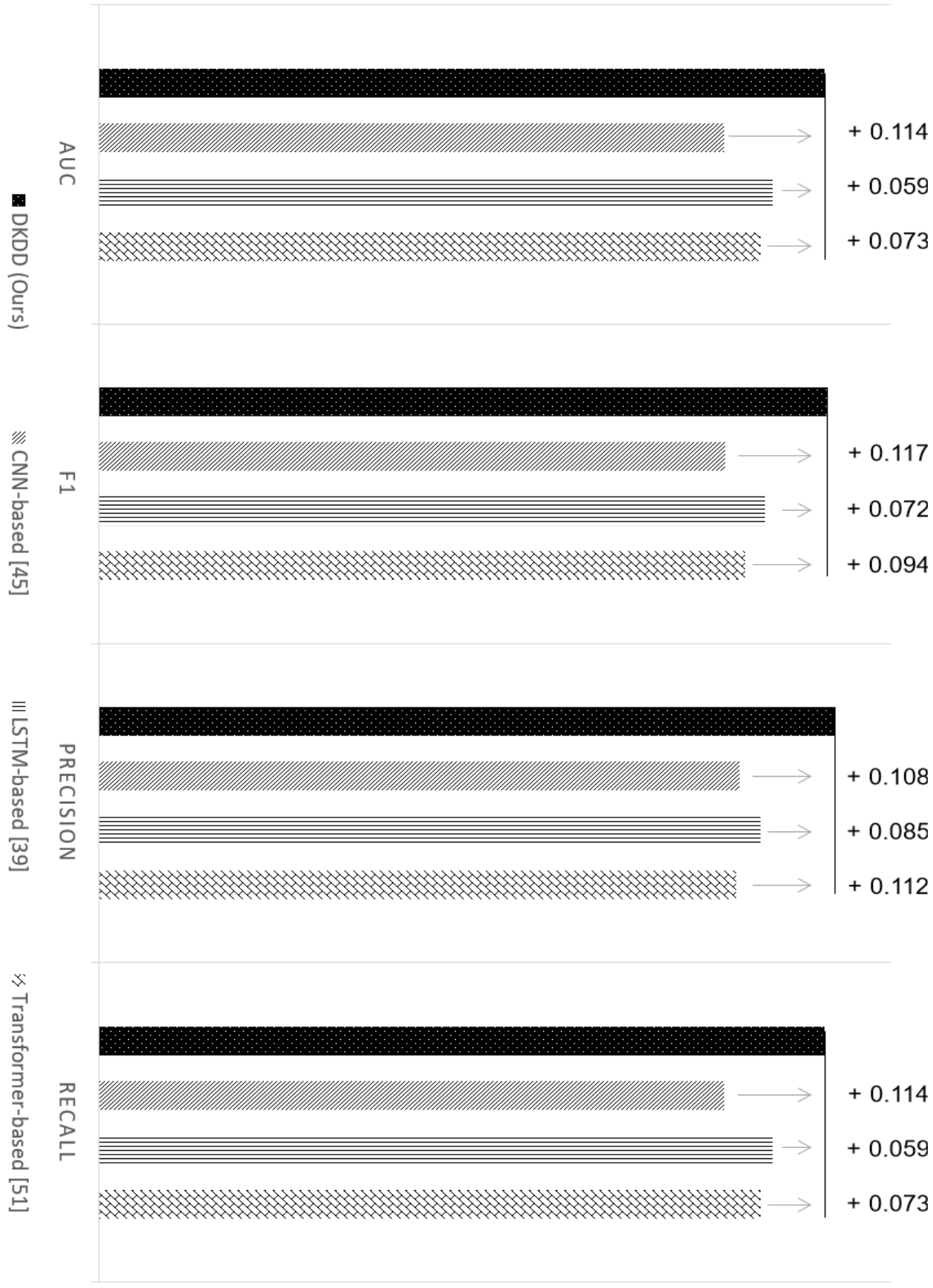

**Figure 6. Knowledge-induced Performance Improvement**

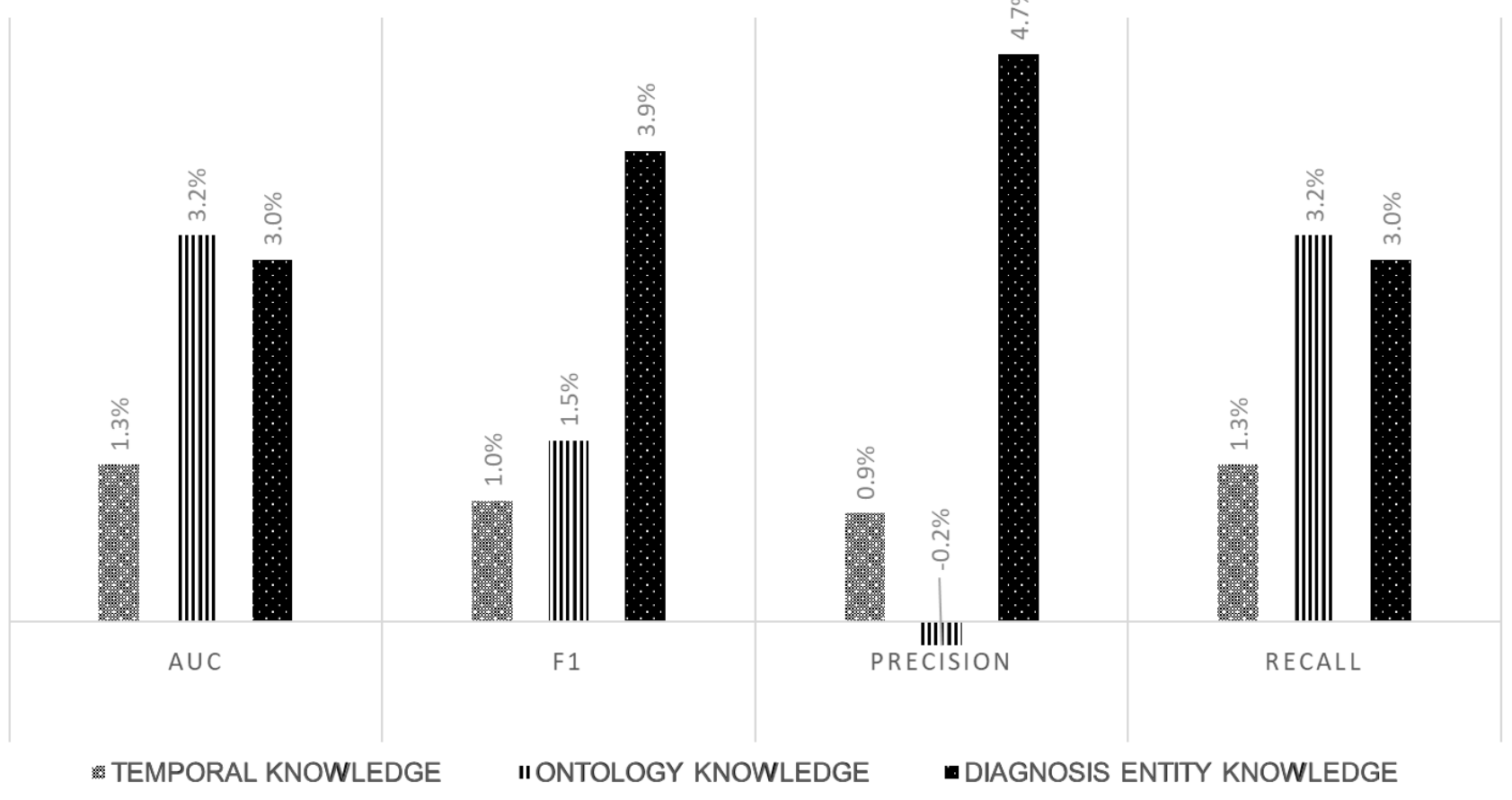

**Figure 7. Explanation of Depression Detection**

| Subject 1 (Digital trace: symptoms, major life events) | Subject 2 (Digital trace: symptoms, life events, treatments) |
| --- | --- |

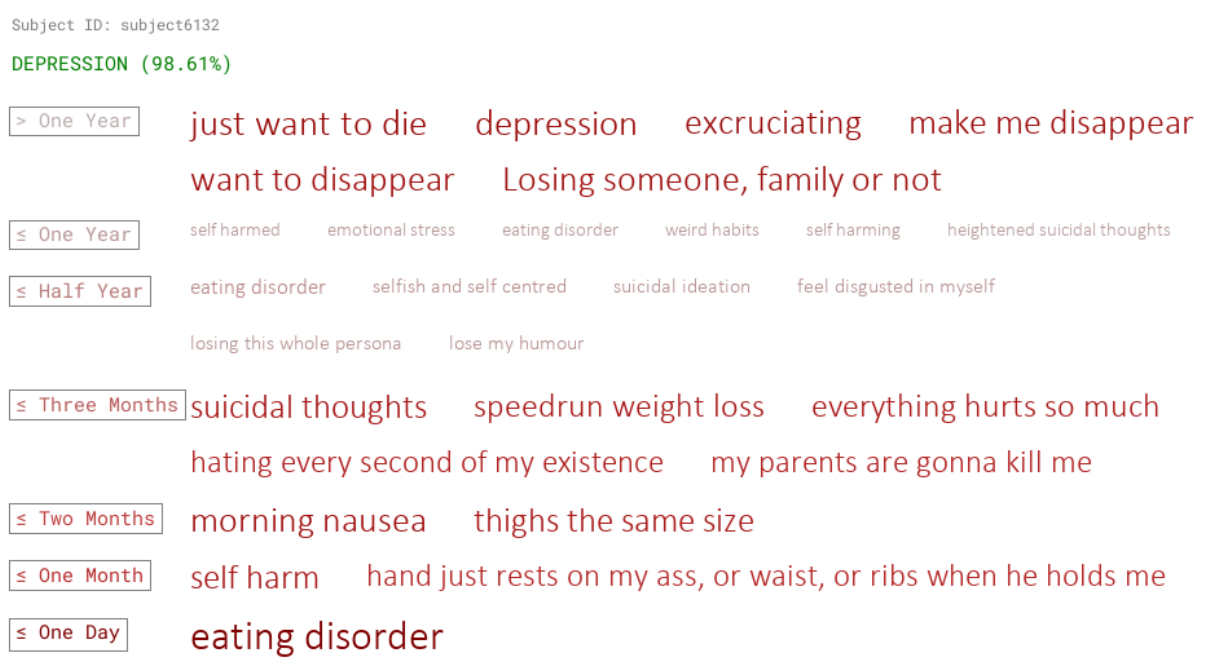

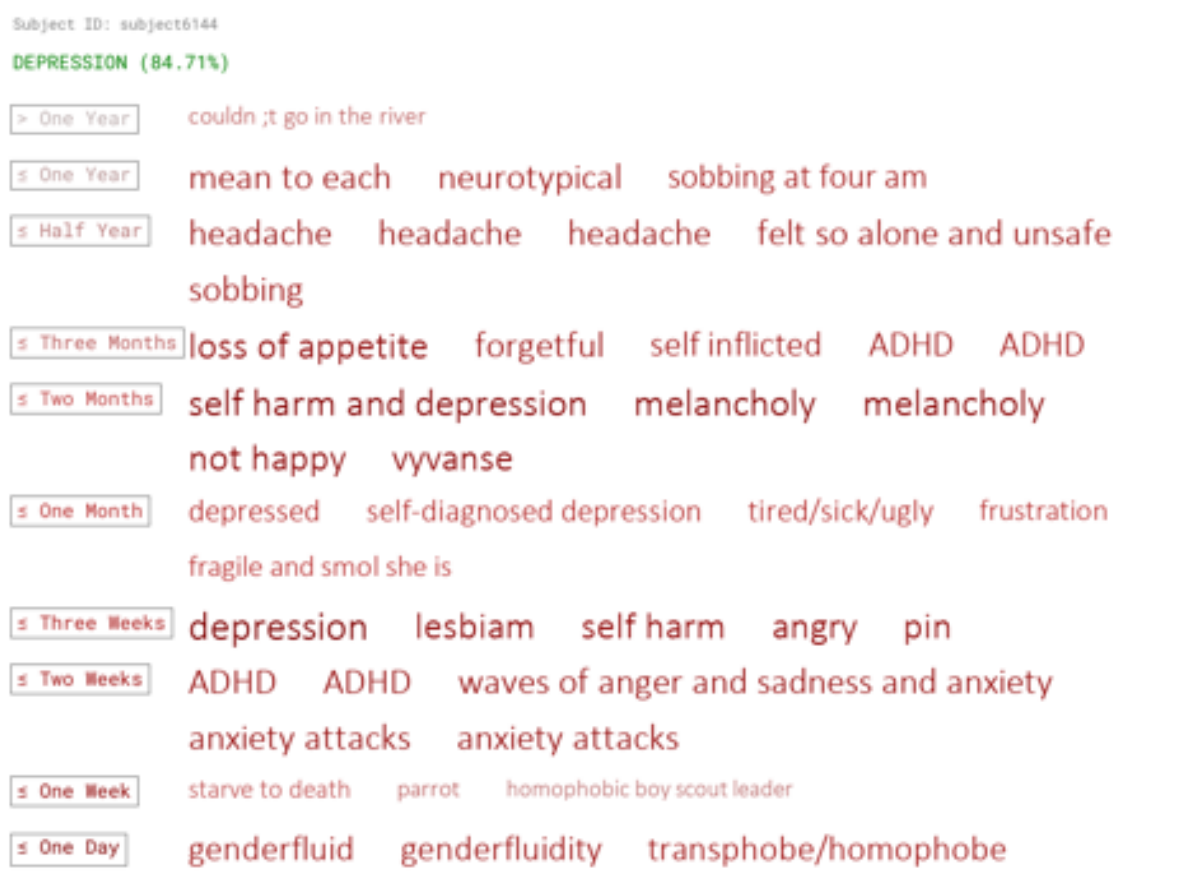

Subject 3 (Digital trace: symptoms, treatments)

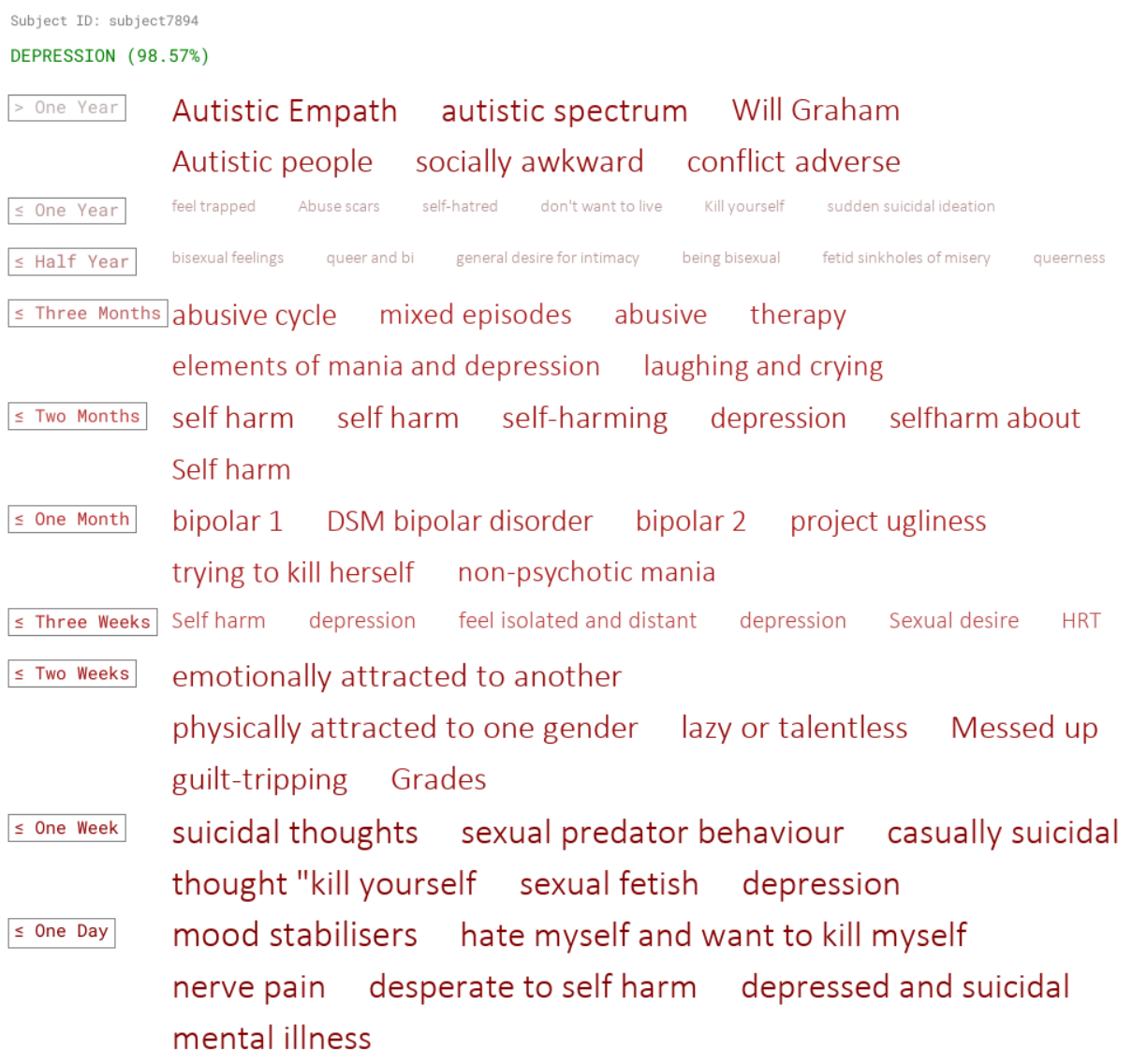

Subject 4 (Digital trace: major life events)

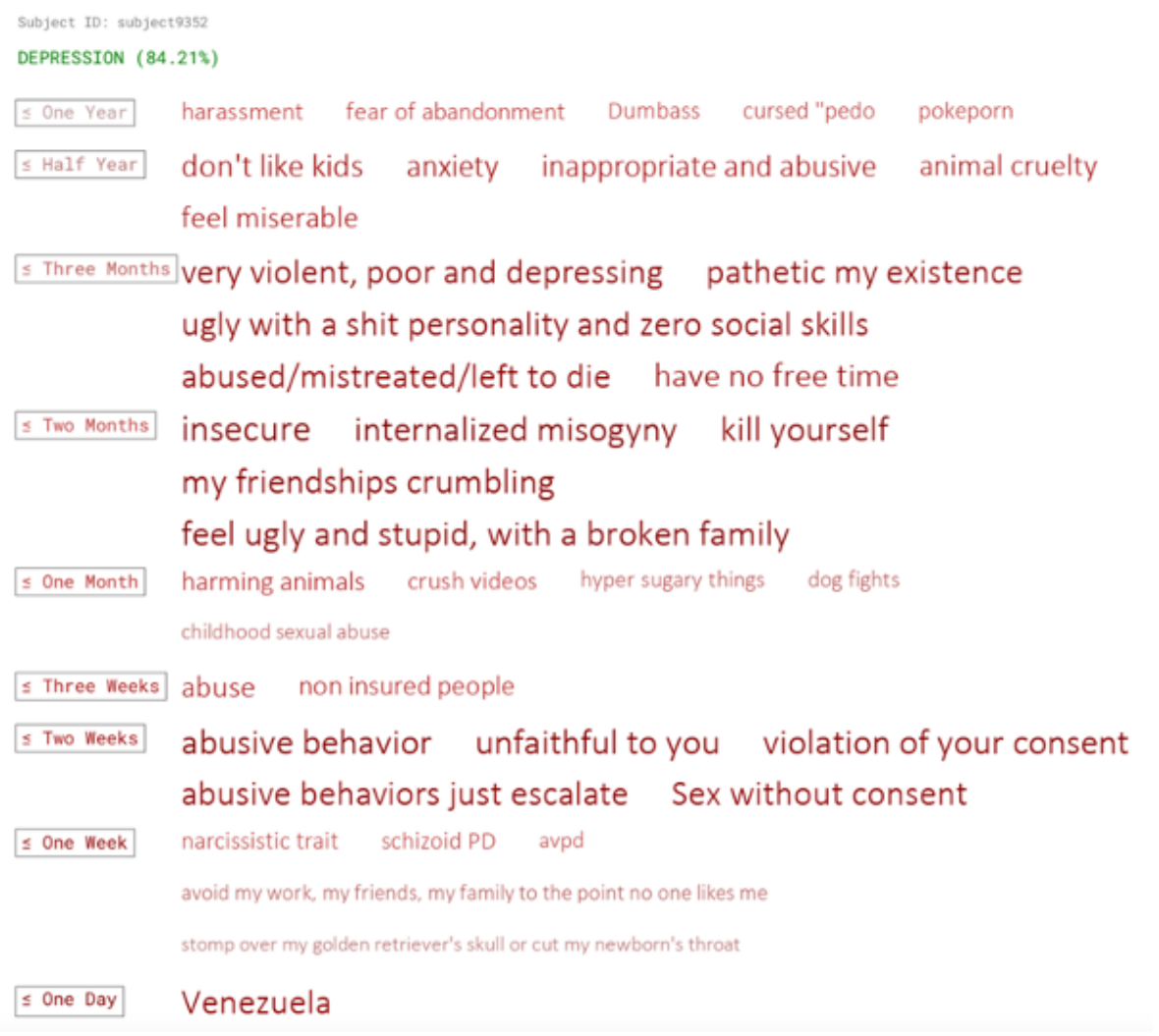

# Depression Detection Using Digital Traces on Social Media: A Knowledge-aware Deep Learning Approach

## Online Appendix

### Depression Ontology Evaluation

In this work, we construct a depression ontology as the medical knowledge base of our proposed DKDD model. To evaluate the quality of the ontology model, we adapt previous work in ontology evaluation [3] and measure the coverage of the ontology by comparing the number of concepts in the ontology with regard to multiple widely used depression diagnosis scales, including DSM-5-TR Self-Rated Level 1 Cross-Cutting Symptom Measure - Adult (DSM-5-TR) [1], Patient Health Questionnaire (PHQ-9) [2], and Quick Inventory of Depressive Symptomatology-Self-Report (QIDS-SR) [4].

Specifically, we define $C = \{c_1, c_2, \ldots c_i, \ldots c_n\}$ as the set of $n$ concepts in the depression ontology $O$, which includes depression symptoms, major life events, and treatments. Let $T = \{t_1, t_2, \ldots t_j, \ldots, t_m\}$ be the set of $m$ medical terminology in the depression diagnosis criteria $D$. The coverage of $O$ to $D$ is calculated as $\frac{\sum_{c \in c(O)} \sum_{t \in T} I(c,t)}{m}$; if there is a $t_j$ or $t_j'$ synonyms in $\{c_1, c_2, \ldots, c_n\}$, set $I(c, t) = 1$, otherwise, set $I(c, t) = 0$. The resulting coverages of the depression ontology to DSM-5-TR, PHQ-9, and QIDS-SR are presented in Table A1.

**Table A1. The Coverage Rate of Depression Ontology to Depression Diagnosis Scales**

| Depression diagnosis scales | DSM-5-TR | PHQ-9 | QIDS-SR |
|---|---|---|---|
| Coverage rate | 85.6% | 95.2% | 93.8% |

The medical terminologies such as "angry," "little energy," and "involvement" are not included in the depression ontology $O$. These terminologies are commonly used in populations other than depression, and therefore, they are not suitable for determining whether a person is depressed outside the context of using the depression diagnosis scales. Overall, the coverage calculation results demonstrate that our ontology can comprehensively cover the widely used depression scales, indicating that it is suitable as the knowledge base for the proposed DKDD model.